\documentclass{article}
\usepackage{slt/spconf,amsmath,graphicx}

\usepackage{times}

\usepackage{amsmath,amsfonts,bm}

\def\eqref#1{equation~\ref{#1}}

\def\1{\bm{1}}

\DeclareMathAlphabet{\mathsfit}{\encodingdefault}{\sfdefault}{m}{sl}
\SetMathAlphabet{\mathsfit}{bold}{\encodingdefault}{\sfdefault}{bx}{n}

\usepackage{hyperref}
\usepackage{url}
\usepackage{graphicx}
\usepackage{amsmath,amsfonts,amssymb}
\usepackage{tabularx, multicol, multirow, booktabs, CJKutf8}

\title{The 2nd FutureDial Challenge: Dialog Systems with Retrieval Augmented Generation (FutureDial-RAG)}

\name{Yucheng Cai$^{1}$,  Si Chen$^{2}$, Yuxuan Wu$^{1}$,  Yi Huang$^{2}$, Junlan Feng$^{*,2}$, 
Zhijian Ou$^{*,1}$\thanks{$^{*}$ Corresponding author. The code is released at \url{https://github.com/futuredialchallenge/2024-RAG/}. This work is partly supported by the National Science and Technology Major Project (2023ZD0121401)}}

\address{
$^{1}$Speech Processing and Machine Intelligence (SPMI) Lab, Tsinghua University, Beijing, China \\
  $^{2}$China Mobile Research Institute, Beijing, China \\
fengjunlan@chinamobile.com, ozj@tsinghua.edu.cn
}
\begin{document}

\maketitle

\begin{abstract}
\end{abstract}
Recently, increasing research interests have focused on retrieval augmented generation (RAG) to mitigate hallucination for large language models (LLMs). 
Following this trend, we launch the FutureDial-RAG challenge at SLT 2024, which aims
at promoting the study of RAG for dialog systems. The challenge builds upon the
MobileCS2 dataset, a real-life customer service datasets with nearly 3000 high-quality dialogs containing annotations for knowledge base query and corresponding results.
Over the dataset, we define two tasks, track 1 for knowledge retrieval and track 2 for response generation, which are core research questions in dialog systems with RAG. We build baseline systems for the two tracks and design metrics to measure whether the systems can perform accurate retrieval and generate informative and coherent response. The baseline results show that it is very challenging to perform well on the two tasks, which encourages the participating teams and the community to study how to make better use of RAG for real-life dialog systems.

\begin{keywords}
Dialog systems, Retrieval augmented generation
\end{keywords}
\vspace{-1.5em}
\section{Introduction}
\vspace{-0.5em}
Developing intelligent dialog systems has been one of the longest running goals in AI. In recent years, significant progress has been made in building dialog systems with the breakthrough of deep learning methods and the large amount of conversational data being made available for system development \cite{budzianowski2018multiwoz,ou2022proceedings,ouyang2022training,achiam2023gpt}. 

There are still 
many challenges toward building future dialog systems. 
The first FutureDial challenge focused on building semi-supervised and reinforced task-oriented dialog systems (FutureDial-SereTOD) \cite{ou2022proceedings,ou2022achallenge}, which was successfully held at EMNLP 2022 SereTOD workshop\footnote{\url{http://seretod.org/}}.
ChatGPT \cite{ouyang2022training}, a newly emerged generative dialog system in the end of 2022, has marked another amazing progress
in engaging users in open-domain dialogs.
However, problems like hallucination and fabrication \cite{alkaissi2023artificial} 
still hinder the usage of such systems in real-life applications like customer service systems, which requires pin-point accuracy. 
Retrieval augmented generation (RAG) \cite{lewis2020retrieval,guu2020realm} has been introduced to enhance dialog systems with retrieved information from external knowledge bases and has attracted increasing interests.
RAG has been shown to be able to help the question-answering and dialog systems to reply with higher accuracy and factuality, providing more informative and grounded responses \cite{humeau2020poly,izacard2021leveraging,shuster2022blenderbot,glass2022re2g,izacard2022contriever,izacard2022atlas,shuster2022language,cai2023knowledge}. 
However, there remain challenges for dialog systems with RAG such as designing retrievers that can retrieve knowledge from multiple knowledge sources, building dialog systems with RAG that can effectively utilize available tools and API-calls for retrieval \cite{schick2023toolformer,yao2023react}, and etc. 
These problems remain under-studied, mainly due to the lack of appropriate dialog datasets to study them.


To further promote the study of how to empower dialog systems with RAG, 
we release a new dataset, called MobileCS2 (Mobile Customer Service) that aims to benchmark and stimulate related researches in this area. 
The dataset is annotated from the real-life customer-service logs instead of the Wizard-of-Oz methods \cite{budzianowski2018multiwoz}. 
Relevant knowledge bases and ground truth retrieved results are annotated so that dialog systems with RAG can be trained over such data.
Note that successful fulfilling of customer-service usually needs to call specialized domain knowledge and/or APIs to retrieve relevant information.
This dataset is very suited to benchmark dialog systems with RAG.
In MobileCS2, there are multiple types of knowledge bases, like user profile, product information and FAQ manual, which bring challenge to the retrieval task in RAG.
Moreover, the dataset contains 
around 3,000 
sessions of unlabeled dialogs along with the same amount of 
sessions of labeled dialogs, which facilitates the study for semi-supervised dialog systems with RAG \cite{zhang-2020-labes,liu2022mga,cai2022advancing,liu2023variational,cai2023knowledge}. We also translate the dataset to make it available in both Chinese and English.

Upon the MobileCS2 dataset, we build a baseline dialog system with RAG. The baseline system consists of a retriever to retrieve relevant knowledge pieces and a generator to generate responses. To comprehensively evaluate the whole system, we design several metrics. For the retriever, we use the recall metrics to evaluate whether the retriever can accurately retrieve the relevant knowledge. For the generator, we assess both the coherency and informativeness of the system and calculate a combined score. 

\begin{figure}[htbp]
\centering
\includegraphics[width=\columnwidth]{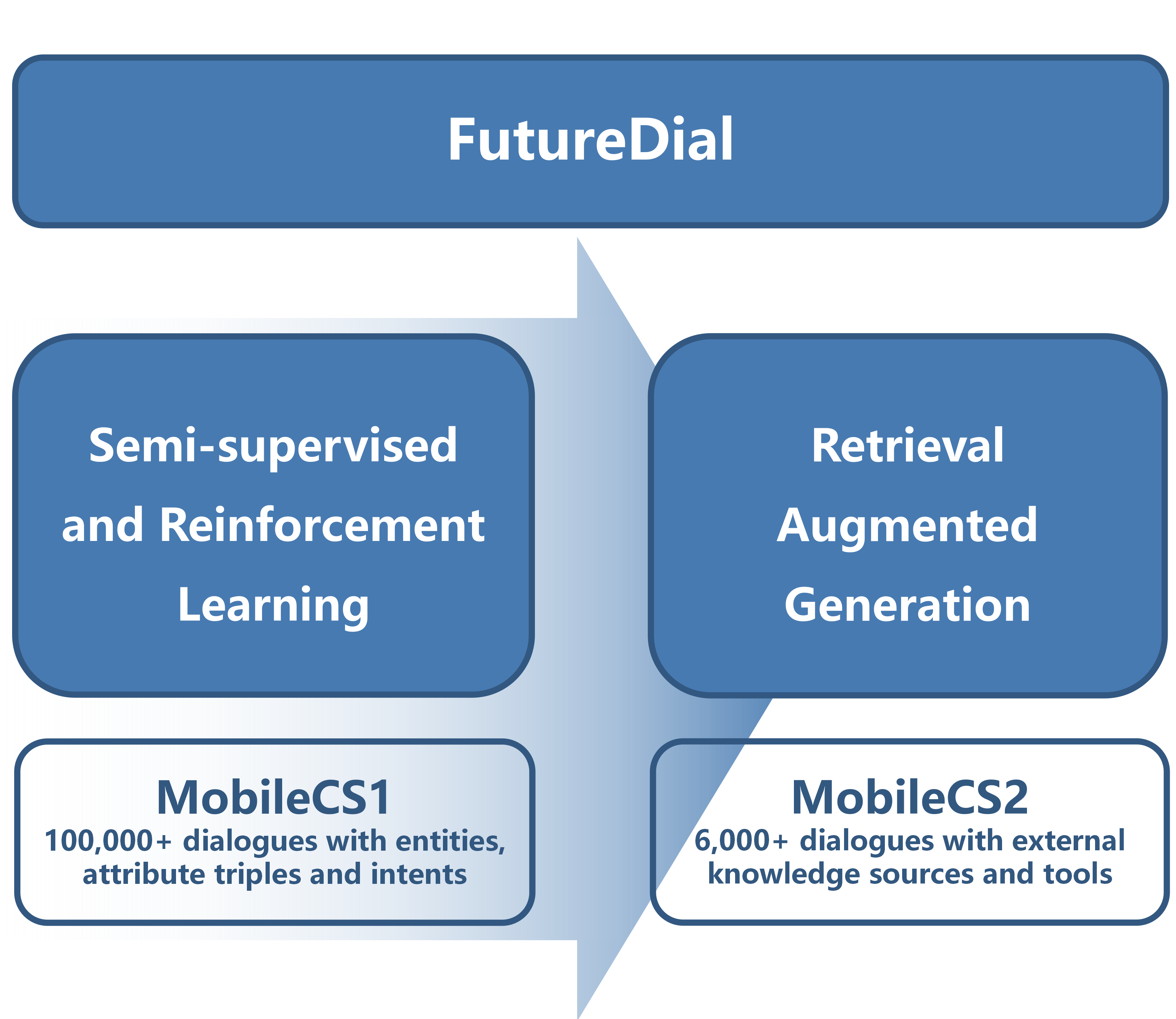}
\vspace{-1.0em}
\caption{Overview of the FutureDial-RAG Challenge: Dialog Systems with Retrieval Augmented Generation.}
\vspace{-0.5em}
\label{fig1}
\end{figure}

Following the success of the 1st FutureDial challenge, the 2nd FutureDial challenge\footnote{\url{http://futuredial.org/}}, co-located with SLT 2024, aims to benchmark and stimulate research in building dialog systems with RAG. With the newly released 
dialog dataset, MobileCS2, the features of the 2nd FutureDial challenge are demonstrated in Figure \ref{fig1}.
We aim to create a forum to discuss key challenges in the field and share findings from real-world applications.

In summary, the main contributions of this work are three folds:

\begin{itemize}
\vspace{-0.5em}
\item We propose the 2nd FutureDial Challenge (FutureDial-RAG) to promote the study of retrieval augmented generation for dialog systems.
\vspace{-0.5em}
\item To support the challenge, we carefully selected and annotated dialogs from real-life customer-service logs and release the MobileCS2 dataset. The dataset is the first dialog dataset for RAG challenge, containing both Chinese and English version.
\vspace{-0.5em}
\item We release a dialog system with RAG as the baseline for the challenge,  as well as metrics and evaluation scripts that can comprehensively evaluate the system. The evaluation results show that the dataset is challenging, which can stimulate research in building dialog systems with RAG. 
\end{itemize}
\vspace{-0.5em}
\section{Related Work}
\vspace{-0.5em}
\subsection{Dialog Datasets with Knowledge Bases}
\vspace{-0.5em}
Building dialog datasets with knowledge bases is challenging, especially those which require annotations for the entities from the knowledge bases for each dialog utterance. The noisy utterances make it hard for the annotators to link the entity mentioned in a dialog to the existing one in the knowledge bases. To mitigate the difficulty of annotating dialog with knowledge bases (especially those with noisy utterances), most previous datasets were built by using Wizard-of-Oz simulated games \cite{wen2017a,budzianowski2018multiwoz,zhu2020crosswoz} to generate annotated dialogs. Instead of annotating real-life dialogs with knowledge bases, the annotators are asked to reconstruct dialogs based on sampled knowledge bases results. Though those datasets have the advantage of high-quality entity annotations with few mistakes, they suffer from the serious problems that the dialogs in the datasets are different from real-life dialog logs, which contain noisy utterances. Dialog systems trained on those datasets are sensitive to noises \cite{liu2022information}, which hinder the usage of those systems for real-life applications.
 
To relieve this issue, the previous work \cite{liu2022information} has proposed to annotate entities from real-life dialogs instead of reconstructing them from the sampled entities. Following the success of \cite{liu2022information}, we also annotate dialogs from real-life customer-service logs. However, in the previous work \cite{liu2022information}, the dialog only contain local knowledge bases aggregated from the annotations of a certain dialog, which is still not so realistic for real-life scenarios. To address this problem, we propose to aggregate the knowledge bases from the annotations of the whole datasets according to different search intents, which will be detailed in Section \ref{sec:dataset}. This setting also support the study of dialog systems with RAG with multiple sources of knowledge bases, which is important for real-life applications of such dialog systems.

\vspace{-0.5em}
\subsection{Retrieval Augmented Generation (RAG)}
\vspace{-0.5em}
Recent researches such as RAG \cite{lewis2020retrieval} and REALM \cite{guu2020realm} have introduced knowledge retrieval models into conditional generation, which greatly improves the quality of generated responses in knowledge-intensive tasks such as open-domain question answering and knowledge-grounded dialog systems. Those works use BM25 or neural network based retrievers such as DPR \cite{karpukhin-etal-2020-dense} to retrieve relevant knowledge given context and generate the answer using the retrieved knowledge pieces. There are several recent studies that improve over the original retrieval-augmented generation systems. Poly-encoder \cite{humeau2020poly} proposes to use a poly encoder instead of the original dual encoder to improve the retrieval accuracy.
Fusion-in-Decoder \cite{izacard2021leveraging} proposes to process retrieved passages independently in the encoder, but jointly fused them in the decoder, which improves the generation quality. 
RetroMAE \cite{xiao2022retromae} proposed to pretrain the retriever with mask auto-encoding objective function to better capture the information of the whole sentence.

However, there still remain several problems that previous works do not comprehensively study due to lack of appropriate datasets. First, to our knowledge, there are no previous works that study how to retrieve from multiple types of knowledge bases, including, for example, user profile (locally to one dialog), product information and FAQ manual (globally in a task).
Second, most RAG-related datasets do not contain annotations of multiple relevant knowledge for a single turn of dialog. These two problems are, however, commonly encountered in real-life scenarios.
In this challenge, the dataset, which is directly annotated from real-life customer-service logs, has the characteristic that helps to study the two problems mentioned above. The challenge provides new challenging settings for researchers to study dialog systems with RAG.
\vspace{-0.5em}
\section{Challenge Overview} 
\vspace{-0.5em}

The 1st FutureDial challenge at EMNLP 2022 \cite{ou2022proceedings,ou2022achallenge,liu2022information,huang2022cmcc}
focused on building semi-supervised and reinforced task-oriented dialog systems (FutureDial-SereTOD), and released a large-scale human-human dialog dataset MobileCS1 (Mobile Customer Service). Following its success, the 2nd FutureDial challenge focuses on building dialog systems with RAG, with the following features:

\begin{itemize}
\vspace{-0.5em}
    \item We release a new dataset from the 
    mobile customer-service logs (MobileCS2) that contains both labeled and unlabeled data, which encourages the study of semi-supervised dialog systems with RAG.
    \vspace{-0.5em}
    \item The dataset enables the study of building dialog systems with knowledge base queries and API calls.
    \vspace{-0.5em}
    \item The dataset is available in both \emph{Chinese} and \emph{English} versions to the public, so that researchers around the world can experiment with this dataset.
\end{itemize}

To enable a dialog system with RAG to provide appropriate answers and services to users, it is essential for the system to utilize knowledge relevant to the conversation context. Therefore the 2nd challenge examines
how dialog systems can retrieve the most appropriate knowledge pieces  from the knowledge base and generate grounded and faithful response to user requests, with the newly released knowledge-grounded dialog dataset, MobileCS2.
The information needed should be retrieved from a given database or API call, which returns specific feedback closely related to real customer service scenarios, such as bill inquiry and package change. Accordingly, the following two tracks are proposed, which are related to the information retrieval of dialog data and the construction of 
dialog systems with RAG in the customer service scenario respectively:
\begin{itemize}
\vspace{-0.5em}
    \item Track 1: Information retrieval based on knowledge bases and dialog context
    \vspace{-0.5em}
    \item Track 2: Dialog systems with retrieval augmented generation
\end{itemize}

Given the context in a dialog, the most relevant knowledge snippet in the multi-source databases should be retrieved by a retrieval model.
So Track 1 aims to build the retrieval model for the dialog system.
Based on retrieved knowledge, Track 2 aims to build a 
retrieval-augmented dialog system in the customer service scenario. The system should generate informative responses leveraging the retrieved results. 

\vspace{-1.0em}
\section{The MobileCS2 Dataset}
\vspace{-0.5em}
\label{sec:dataset}

\begin{table}[t] \small
    \caption{Statistics of the MobileCS2 dataset. Knowledge-retrieval-needed denotes those turns that api\_query$\neq$NULL.}
	\centering
	\resizebox{1\linewidth}{!}{
			\begin{tabular}{ c c c c c}
				\toprule
				&Train  &Dev &Test & Total\\ 
				\midrule
    Dialogs & 1926     & 412&  413&2751\\
    Turns & 16120    & 3246&  3240& 22606\\
  Knowledge-retrieval-needed & 4314     & 808&  817&5939\\

\bottomrule
		\end{tabular}}
  \vspace{-1.0em}
		\label{tab:data-stat}
\end{table}

To support the 2nd FutureDial challenge and the study of dialog systems with RAG, we release the MobileCS2 dataset. 
The MobileCS2 dataset is derived from the real-world mobile conversational scenarios and comprises around 6,000 processed dialog logs (nearly 3,000 carefully annotated) between customers and customer service staffs. The statistics of the MobileCS2 dataset is shown in Table \ref{tab:data-stat}. It can serve for research aims such as the development of conversational models, colloquial human-to-human dialog systems, and data-driven systematic dialog analysis.

\begin{table*}[h]
 {
     \caption{Detailed description for api\_query annotation. }
     \label{tab:intent collection}
     \begin{tabularx}{\textwidth}{l|l|X} 
            \toprule
            \textbf{Main\_class} & \textbf{api\_query} & \textbf{Description} \\ \midrule
            QA & [QA] & Consult the FAQ manual, which includes a collection of commonly asked questions such as recent promotional packages and general business regulations. \\ \midrule
            NULL & - & Based on the contextual information, customer service personnel can successfully complete the conversation without the need for additional inquiries. \\ \midrule
            {\multirow{3}{*}{API-Inquiry}} & Search for products information & Inquire about the current business information of the mobile company
            , such as specific packages, data plans, etc. \\ \cline{2-3}
            & Search for user information & Inquire about the services that the user currently possesses, including the current package, current monthly fee, and current data usage. \\ \cline{2-3}
            & Search for other information & Inquire about other key information used to complete the dialog. For example, inquiring about text messages regarding excessive data usage alerts sent by the mobile company 
            in the historical trajectory, querying the address of the business hall, etc. \\ \midrule
            API-Cancel & Cancel business& Revoke a certain service currently possessed by the user. \\ \midrule
            API-Handle & Handle business & Process a new service for the user. \\ \midrule
            API-Verification & Verify identity & Send verification codes, passwords, or other related customer service verification operations to the user. \\
            \bottomrule
        \end{tabularx}}
        \vspace{-1.5em}
\end{table*}

\vspace{-1.0em}
\subsection{Annotation Details}
\vspace{-0.5em}
In the customer service scenario, there are some knowledge or information that the customer service agent needs to get from knowledge bases (KBs) in order to correctly respond to the user. 
Therefore, to annotate the necessary knowledge or information, the annotators should imagine themselves as customer service agents. 
When presented with a dialog, annotators are required to identify the agent's intent (labeled by api\_query) at each turn. If the intent is to query the KBs to seek external information and the response contains specific details, the annotator should perform a retrospective analysis based on the information provided in the response and annotate the corresponding query result.
For those turns that knowledge retrieval is not needed by the agent in order to respond to the user, api\_query is labeled to be NULL.
Specifically, Table \ref{tab:intent collection} contains the set of intents and the explanations of each intent, which are provided to annotators for their reference during the annotation process. For example, given the dialog ``Help me check my package'', the annotator needs to identify the intent ``Search for user information'' and then annotate the package that appears in the customer service's response into the query result. 

We recruited 6 customer service staffs for the annotation, which are divided into 2 teams. The annotation is conducted dialog by dialog, and the labeling task for one dialog is assigned to an arbitrary annotator, and the annotation process takes about a week.
To ensure the quality of the dataset, cross-validation is conducted between the 2 teams, and 100 annotated dialogs are checked by the other team everyday. The cross-validation agreement rate is 97 percent, which shows the dataset is of high quality. After annotation, the dataset is desensitized to remove sensitvie personal information like individual names, ids, and phone numbers. 

\begin{figure*}[h]
    \centering
    \includegraphics[width=\linewidth]{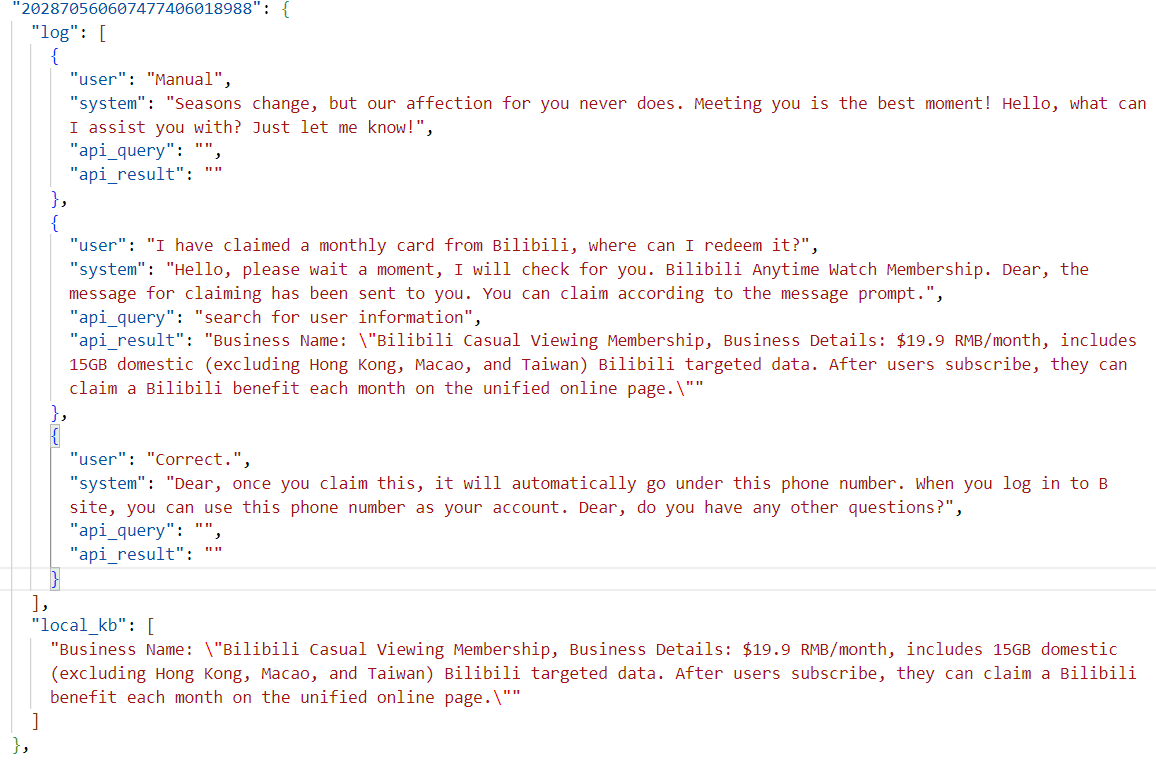}
    \vspace{-1.5em}
    \caption{An example of annotated dialogs. }
    \vspace{-1.5em}
    \label{fig:Annotation English Sample}
\end{figure*}

In the final dataset, each sample in the dataset represents a dialog. At each turn of the dialog, there are two types of information to be annotated: customer service intent (api\_query) and customer service query results (api\_result). An example of the annotated data dialog data is shown in Figure \ref{fig:Annotation English Sample}.

\vspace{-0.5em}
\subsection{Post-processing}
\vspace{-0.5em}
Based on the annotation data, it is possible to aggregate the information in the dataset and simulate the information that the agents can access in real-world services. For turns annotated with the inquiry [QA], the information can be aggregated into an FAQ (Frequently Asked Questions) handbook across the entire dataset ($KB_{FAQ}$). Turns labeled as ``Search for user information'' can be consolidated into a user database ($KB_{user}$) within a single dialog. Meanwhile, turns labeled as ``search for products information'' can be aggregated into a product database ($KB_{product}$) across the entire dataset. These three databases largely emulate the channels through which the agents acquire knowledge in real-world settings.
In MobileCS2, the size of $KB_{FAQ}\cup KB_{product}$ is 4108, and the average size of $KB_{user}$ is 0.535.

To make the dataset more widely accessible, we translate the original Chinese dataset into English using the ChatGPT API. It is important to ensure that the annotated knowledge entities are consistent in the translated dialogs and the knowledge annotations. We first translate the knowledge annotations, and substitute the translated knowledge annotations into the dialog corpus. The English-Chinese hybrid dialogs are then sent to ChatGPT and translated into English. We find that ChatGPT can translate such hybrid data well, meanwhile maintaining the annotated knowledge entities consistent with the translated annotations.

\vspace{-0.5em}
\section{The Baseline System}
\vspace{-0.5em}

\begin{figure}[htbp]
\centering
\includegraphics[width=\columnwidth]{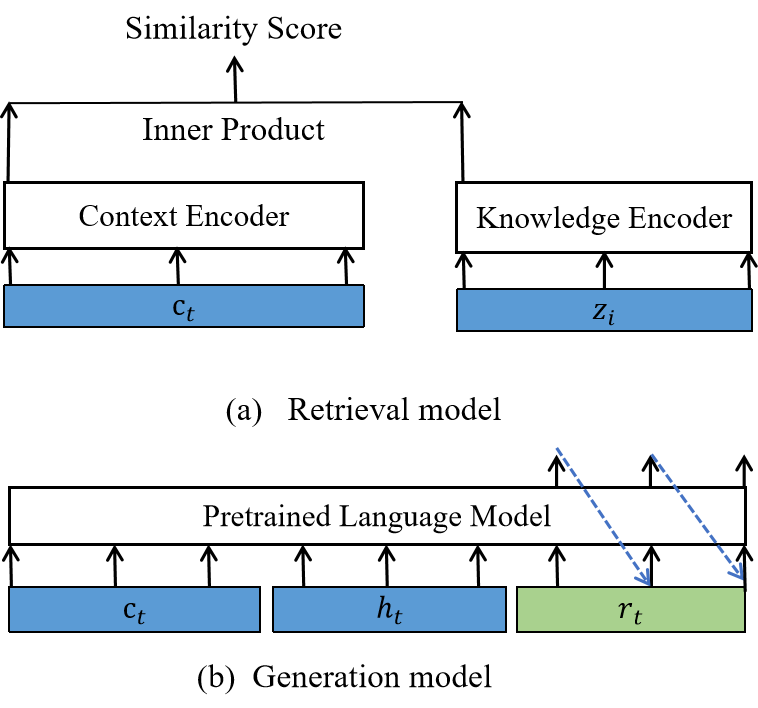}
\vspace{-1.5em}
\caption{Overview of our baseline systems:  (a) the retrieval model, (b) the generation model. }
\vspace{-0.5em}
\label{fig:model}
\end{figure}

We use RAG-based \cite{lewis2020retrieval,cai2023knowledge} methods to build our baseline system.
Dialog systems with RAG aim to retrieve relevant knowledge pieces given the dialog context and generate system response using the retrieved knowledge. We take into consideration multiple important characteristics of the MobileCS2 dataset to design the baseline system, such as adding the unique user profile for each user to the knowledge base and considering multiple relevant knowledge pieces given context. This makes the baseline system over the MobilleCS2 dataset more suitable for real-life scenarios, which is different from prior work in knowledge grounded dialog systems  \cite{lewis2020retrieval,cai2023knowledge}.

To introduce the baseline dialog system with RAG on MobileCS2, we make the following definitions. Assume we have a dialog $X$ with $T$ turns of user utterances and system responses, denoted by $u_{1},r_{1},\cdots,u_{T},r_{T}$ respectively. For each dialog, we assume that there is a knowledge base that is necessary for the system to respond correctly. In  MobileCS2, the knowledge base is made up of the user information, which is unique for each dialog, the product information list, and the FAQ list for commonly asked questions. Therefore, for the dialog $X$, the knowledge base $KB_{X}$ can be denoted as: 
$KB_{X} \triangleq KB_{user}\cup KB_{FAQ}\cup KB_{product}$.

At turn $t$ of a dialog $X$, based on dialog context $c_{t} \triangleq u_{1} \oplus r_{1} \oplus \cdots \oplus u_{t-1} \oplus r_{t-1} 
\oplus  u_{t}$ ($\oplus$ means sequence concatenation) and the knowledge base $KB_{X}$, the system uses a retriever, which is shown in Figure \ref{fig:model}(a), 
to get the relevant knowledge $h_{t}$ from the knowledge base and generates appropriate responses with the generator $p_\theta(r_{t} \mid c_{t},h_{t})$, which is shown in Figure \ref{fig:model}(b). 

The retrieval model is implemented with the dual-encoder architecture, as shown in Figure \ref{fig:model}(a). To train the retrieval model, 
we consider each knowledge piece $z_i~(i=1, 2, \cdots, K)$ in $KB_{X}$ and model the retrieval distribution of $p_\eta(z_{i}  \mid c_{t})$ as in   \cite{lewis2020retrieval}: 
\vspace{-0.5em}
\begin{align}
p_\eta(z_{i} \mid c_{t}) \propto \exp \left(\operatorname{Encoder}_p(z_{i})^{\top} \operatorname{Encoder}_c (c_{t})\right) 
\end{align}
$\operatorname{Encoder}_p$ and $\operatorname{Encoder}_c $ are both initialized with a BERT-based pretrained model \cite{devlin2019bert}. The probability is optimized with the standard cross entropy loss, with the positive pieces 
$z \in Z_{+}$ labeled in the dataset:
\begin{equation}
\mathcal{L}_{ret}
=-\frac{1}{\mid Z_{+} \mid}\sum_{z \in Z_{+}} \log \frac{p_\eta (z \mid c_{t})}{ p_\eta (z \mid c_{t}) +\sum_{i=1, z_{i}\neq z}^K  p_\eta (z_{i} \mid c_{t})}
\label{eq:retriever}
\end{equation}

The knowledge piece encoder $\operatorname{Encoder}_p$ is fixed during the training, and the context encoder $\operatorname{Encoder}_c $ is trained with the loss in Eq. \ref{eq:retriever}, following the setting in \cite{karpukhin-etal-2020-dense}.

To train the dialog system $p_\theta(r_{t} \mid c_{t},h_{t})$, we use the standard auto-regressive loss to optimize the generation probability :
\vspace{-1.0em}
\begin{align}
{p}_{\theta}(r_{t} \mid c_{t},h_{t})  
=\prod_{l=1}^{|r_{t}|}
p_{\theta}(y^l \mid c_{t}, h_{t}, y^1, \ldots, y^{l-1}) 
\label{eq:gpt2}
\end{align}
where $|\cdot|$ denotes the length in tokens, and $y^l$ the $l$-th token of $r_t$ and $p_\theta$ is initialized with a GPT-based pretrained lanugae model \cite{radford2019gpt2}.
\vspace{-0.5em}
\section{evaluation}
\vspace{-0.5em}
Given a dialog $X$ and its knowledge base $KB_{X}$, the retrieval system needs to rank the relevance score for each knowledge piece in the $KB_{X}$. 
We use the commonly used recall metrics to assess the retrieval system. To get the recall@k metrics, we calculate whether the ground-truth knowledge piece is in the top-k retrieved knowledge pieces. To comprehensively evaluate the retrieval quality of the system, we calculate the 
sum of the recall for $k=1,5,20$ as the final score: $score_{retriever} = recall@1 + recall@5 + recall@20$. The evaluation results for the baseline system on the Dev set are shown in Table \ref{tab:baseline-1} \footnote{We use the BGE \cite{chen2024bge} model as our backbone. The resulting model is selected based on the Score on the Dev set.}. The results show that it is challenging for the retriever to achieve high recall even with top-20 retrieved documents.  

\begin{table}[t] \tiny
    \caption{Baseline results for the retrieval task (Track1).}
	\centering
	\resizebox{1\linewidth}{!}{
			\begin{tabular}{ c c c c c}
				\toprule
				recall@1  &recall@5 &recall@20 &Score \\ 
				\midrule
    0.225 & 0.387     & 0.573&  1.185\\

\bottomrule
		\end{tabular}}
		\label{tab:baseline-1}
\end{table}

To generate the suitable  system response, relevant knowledge pieces are first retrieved using the retrieval system. Given the retrieved knowledge pieces, the generator can generate response based on the retrieved knowledge. The generated response is evaluated by measuring the similarity score with the ground-truth response (\emph{BLEU} and \emph{BERTScore}) and whether the system correctly provides the requested information by the user (\emph{Inform Rate}).
 \emph{BLEU} is used to measure the fluency of the generated responses by analyzing the amount of n-gram overlap between the real responses and the generated responses. 
  \emph{BERTScore} \cite{zhang2019bertscore} is used to measure the semantic similarity of the generated responses with the oracle responses  by using a pretrained BERT model. 
 \emph{Inform Rate} refers to how often the system response is able to cover the requested information by the user. 
 The final score of the generator is computed as 
$score_{generator} = 0.5*(BLEU/100 + BERTScore) + Inform$. The evaluation results for the baseline system are shown in Table \ref{tab:baseline-2} \footnote{We use the Chinese GPT-2 \cite{radford2019gpt2} model as our backbone. The resulting model is selected based on the Score on the Dev set.}. The $Inform$ of the results is relatively low, presumably due to the low recall@1 score of the retriever and the cascaded error of the whole system. This shows the difficulty to build a RAG system that can accurately provide required knowledge in real-world scenarios.

\begin{table}[t] \tiny
    \caption{Baseline results for the response generation task (Track2).}
	\centering
	\resizebox{1\linewidth}{!}{
			\begin{tabular}{ c c c c c}
				\toprule
				BLEU-4  &BERTScore &Inform &Score \\ 
				\midrule
    14.54 & 0.639     & 0.092 &  0.484\\

\bottomrule
		\end{tabular}}
		\vspace{-0.5em}\label{tab:baseline-2}
\end{table}

\vspace{-1.0em}
\section{Challenge rules}
\vspace{-1.0em}
Each team should submit the results for the challenge via  
the official email before the Entry Submission Deadline. 
The submission to the challenge is required to contain a system executable with all the dependencies or an encapsulated web service to be called, with a clear README documentation for running the system over the evaluation data.
In either manners, the system’s processing speed should be no less than 10 tokens per second.
The submission should also provide a System Description Document (SDD), introducing the submitted system.
The organizers of the challenge either run the systems over a server with Nvidia A100*4 hardware, or call the web service to evaluate the submitted systems and obtain the final scores. 

\vspace{-0.5em}
\section{Conclusion}
\vspace{-0.5em}
We present the FutureDial-RAG challenge at SLT 2024, which aims
at promoting the study of RAG for dialog systems. To support the challenge, we build the MobileCS2 dataset, a real-life customer service datasets with over 3000 high-quality dialogs, containing annotations for knowledge base query and corresponding results. 
Upon the dataset, we define two tasks, track 1 for knowledge retrieval and track 2 for response generation, which are core research questions in dialog systems with RAG. We build baseline systems for the two tracks and design metrics to measure whether the system can perform accurate retrieval and generate informative and coherent response. The baseline results show that it is very challenging to perform well on the two tasks. We hope that the dataset and the baseline systems will be helpful to foster the community to study how to make better use of RAG for dialog systems.
\bibliographystyle{slt/IEEEbib}
\bibliography{reference}

\newpage
\appendix

\end{document}